\pdfoutput=1

\documentclass[11pt]{article}

\usepackage{EMNLP2022}

\usepackage{times}
\usepackage{latexsym}

\usepackage[T1]{fontenc}

\usepackage[utf8]{inputenc}

\usepackage{microtype}

\usepackage{inconsolata}

\usepackage{graphicx}
\usepackage{amsmath,amssymb} 
\usepackage{color}
\usepackage{epsfig}
\usepackage{tabularx}
\usepackage{booktabs}
\usepackage{multirow}
\usepackage{array}
\usepackage{xspace}
\usepackage{balance}
\usepackage{enumitem}
\usepackage{mathtools}
\usepackage{rotating}
\usepackage{tabulary}
\usepackage[export]{adjustbox}
\usepackage{tikz,ctable}
\usepackage{diagbox}
\usepackage{subcaption}
\usepackage[linesnumbered,ruled]{algorithm2e}
\usepackage{bbm}
\usepackage{wrapfig}
\usepackage{todonotes}

\usepackage{verbatim}

\title{A Neural-Symbolic Approach to Natural Language Understanding}

\author{Zhixuan Liu$^*$ \\ Shanghai Jiaotong University \\ \texttt{lzx993124494@sjtu.edu.cn}
\And Zihao Wang$^*$ \\ CSE, HKUST \\ \texttt{zwanggc@cse.ust.hk}
\AND Yuan Lin \\ ByteDance AI Lab \\ \texttt{linyuan.0@bytedance.com}
\And Hang Li \\ ByteDance AI Lab \\ \texttt{lihang.lh@bytedance.com}}

\begin{document}
\maketitle

\def\thefootnote{*}\footnotetext{The work was done when the first and second authors were interns at ByteDance AI Lab.}

\begin{abstract}
Deep neural networks, empowered by pre-trained language models, have achieved remarkable results in natural language understanding (NLU) tasks. 
However, their performances can drastically deteriorate when logical reasoning is needed. This is because NLU in principle depends on not only analogical reasoning, which deep neural networks are good at, but also logical reasoning. According to the dual-process theory, analogical reasoning and logical reasoning are respectively carried out by \emph{System 1} and \emph{System 2} in the human brain. Inspired by the theory, we present a novel framework for NLU called Neural-Symbolic Processor (NSP), which performs analogical reasoning based on neural processing and logical reasoning based on both neural and symbolic processing. As a case study, we conduct experiments on two NLU tasks, question answering (QA) and natural language inference (NLI), when numerical reasoning (a type of logical reasoning) is necessary. The experimental results show that our method significantly outperforms state-of-the-art methods in both tasks.$^1$

\end{abstract}

\def\thefootnote{1}\footnotetext{The code and data are available at \url{https://github.com/chadlzx/NSP_QA}; \url{https://github.com/zihao-wang/Number-NLI}.}

\section{Introduction}

Natural language understanding (NLU) has made remarkable progress recently, when pre-trained language models such as BERT~\cite{devlin2018bert}, RoBERTa~\cite{liu2019roberta}, and BART~\cite{lewis2020bart} are exploited. The deep neural networks based on pre-trained language models even exhibit performances superior to humans in the tasks of question answering (QA)~\cite{rajpurkar2016squad} and natural language inference (NLI)~\cite{ghaeini2018dr}. However, language understanding requires not only analogical reasoning, which deep neural networks are good at~\cite{bengio2021deep}, but also logical reasoning, including numerical reasoning. For example, to answer the questions in Table~\ref{tab:QA_num_examples} or to infer the entailment relations in Table~\ref{tab:NLI_num_examples}, we need to first `interpret' the meanings of the input texts and then perform numerical reasoning, in general, logical reasoning, to obtain the final results. Exploiting deep neural networks alone would not easily accomplish the goal. Recently, there has been research to address the problems in QA and NLI. For example, a dataset called Discrete Reasoning Over Paragraphs (DROP) has been created for QA, and several methods have been proposed~\cite{dua2019drop,ran2019numnet,chen2020question}. Among them, Numerically-Aware QANet (NAQANet)~\cite{dua2019drop} utilizes deep neural networks to individually solve the sub-problems of span extraction, counting, and numerical addition/subtraction. A dataset called AWPNLI~\cite{ravichander2019equate} has also been created for NLI in which numerical reasoning is needed.

\begin{figure*}[htbp]
\setlength{\abovecaptionskip}{0.3cm} 
\setlength{\belowcaptionskip}{-0.5cm}
    \centering
    \includegraphics[scale=0.55, trim=0 0 0 0]{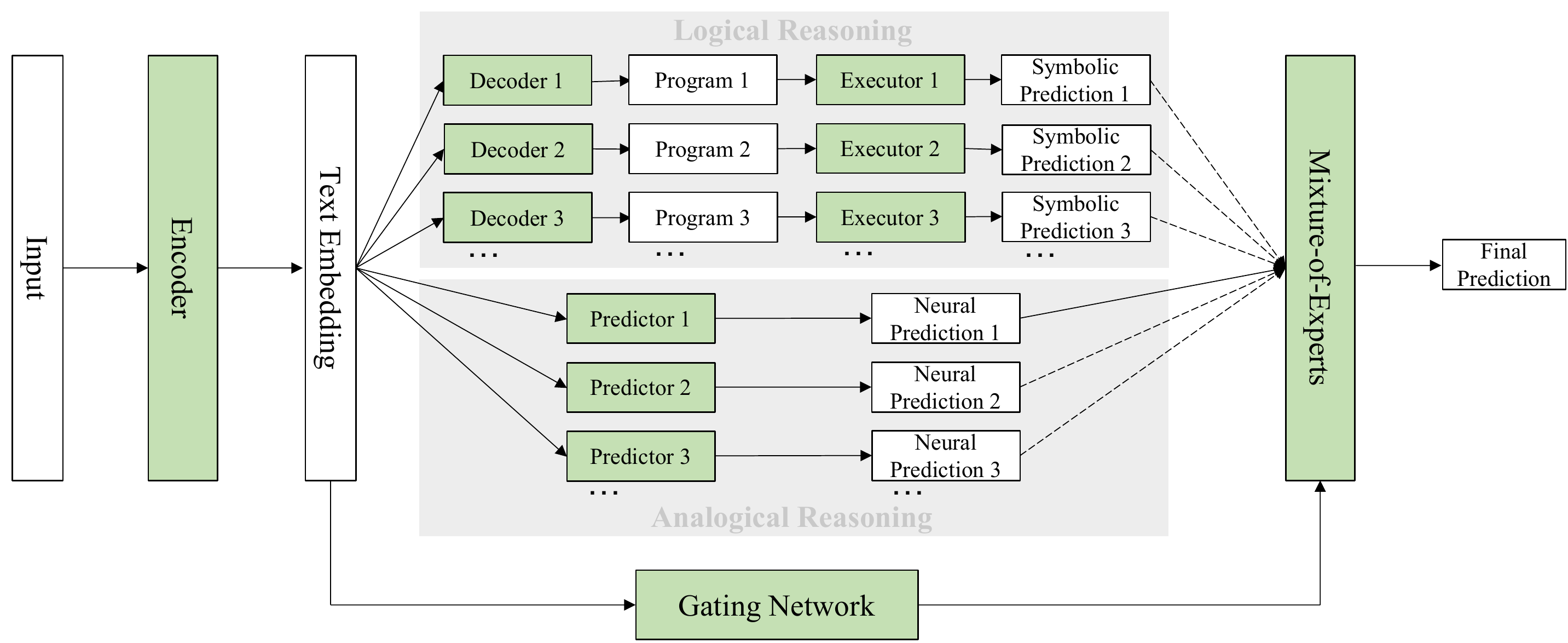}
    \caption{An overview of the Neural-Symbolic Processor framework. Analogical reasoning is performed by the predictors (neural processing). Logical reasoning is performed by the decoders and executors (neural and symbolic processing). A mixture-of-experts is used to make the final prediction.}
    \label{framework}
\end{figure*} 

According to the dual-process theory developed by the psychologist Kahneman and others~\cite{kahneman}, human thinking is carried out by two different systems in the brain. \emph{System 1} is a fast, unconscious, and effortless mode of thinking, often associated with analogical reasoning. In contrast, \emph{System 2} is a slow, conscious, and effortful mode of thinking, also evoking logical reasoning. The theory should also hold for 
 human language understanding, a specific case of thinking. Inspired by the theory, we propose a new framework for language understanding, which performs both analogical reasoning and logical reasoning, corresponding to \emph{System 1} and \emph{System 2} respectively.
 In fact, designing AI systems containing \emph{System 1} and \emph{System 2} is a popular research topic recently (e.g., ~\cite{bengio2021deep}). Our key idea is to employ a neural network to conduct analogical reasoning on the text input as usual and in the meantime, to employ another neural network to translate the text input into a program and a symbolic system to execute the program to perform logical reasoning.

The new framework for natural language understanding, called Neural-Symbolic Processor (NSP), is shown in Figure~\ref{framework}. First, the encoder transforms the input texts (the question and text in QA, the two texts in NLI) into a text embedding (an intermediate representation). Then, the predictors take the text embedding as input and generate neural predictions. In parallel, the decoders transform the text embedding into programs. Note that the encoder and the decoders form sequence-to-sequence models. Moreover, the executors take the programs as input and generate symbolic predictions. Finally, the mixture-of-experts (with a gating network) takes all the neural predictions and symbolic predictions as input and selects one of the predictions to make the final prediction. 

We evaluate the NSP framework in QA and NLI. For the QA task, we sample a subset of DROP (about 32K instances), referred to as DROP-subset, and annotate a program for each question-answer pair requiring numerical reasoning. The experiments on the DROP-susbset show that our approach outperforms the baselines by 2.40\% and 2.51\% in terms of F1 score and exact match. In particular, for question-answer pairs requiring multiplication, division, and averaging, our approach improves F1 by 56.42\%. For question-answer pairs requiring addition and subtraction, our approach improves F1 by 3.91\%. For the NLI task, we use the AWPNLI dataset~\cite{ravichander2019equate} and also annotate programs for each text pair. The experiments on AWPNLI show that our approach exceeds the baseline with a large margin of 20.7\% in terms of F1 score. 

The contribution of our work is as follows:

\begin{itemize}
    \item We propose a new framework for NLU, Neural-Symbolic Processor, to conduct both analogical reasoning and logical reasoning. 
    \item We perform experiments on QA and NLI to verify the effectiveness of our approach and show that our approach can achieve remarkable improvements in the tasks when they need numerical reasoning. 
     \item We add programs into the two datasets of QA and NLI, DROP-subset and AWPNLI, and will release the annotated data. 
\end{itemize}

\begin{table*}[]
\setlength{\belowcaptionskip}{-0.5cm}
    \centering
    \begin{tabular}{p{7.8cm}p{3.5cm}p{3.7cm}}
        \toprule
        Passage & Question & Program \& Prediction \\
        \hline
        
        ... kicker Kris Brown getting a 53 - yard{\color{orange}@N9} and a 24 - yard{\color{orange}@N10} field goal. ... & How many more yards was Kris Browns's first{\color{orange}@Q1} field goal over his second{\color{orange}@Q2}? &  {\bf Program}: diff(N9,N10) \newline {\bf Symbolic Prediction}: \newline 29 \newline {\bf Ground-Truth}: 29 \\\midrule
        
        ... The first{\color{orange}@N1} issue in 1942{\color{orange}@N2} consisted of denominations of 1{\color{orange}@N3}, 5{\color{orange}@N4}, 10{\color{orange}@N5} and 50{\color{orange}@N6} centavos and 1{\color{orange}@N7}, 5{\color{orange}@N8}, and 10{\color{orange}@N9} Pesos. The next year{\color{orange}@N10} brought "replacement notes" of the 1{\color{orange}@N11}, 5{\color{orange}@N12} and 10{\color{orange}@N13} Pesos ... & In which year{\color{orange}@Q1} were there replacement notes of the 1{\color{orange}@Q2} , 5{\color{orange}@Q3} , and 10{\color{orange}@Q4} pesos ? & {\bf Program}: add(N2,N10) \newline {\bf Symbolic Prediction}: \newline 1943 \newline {\bf Ground-Truth}: 1943 \\
        \bottomrule
    \end{tabular}
    \caption{Two examples of question answering requiring numerical reasoning. In our method, the numbers in the input are attached with special tokens (in orange). The programs are generated for logical reasoning, in addition to analogical reasoning. The symbolic predictions are obtained by executions of the programs.}
    \label{tab:QA_num_examples}
\end{table*}

\section{Related Work}

\paragraph{Question Answering with Logical Reasoning}

Given a short text and a question, the QA task is to predict the answer obtained from the short text. 
QA has made considerable progress in recent years. Models such as BiDAF~\cite{seo2016bidirectional}, R-NET~\cite{wang2017r}, and QANet~\cite{yu2018qanet} have been proposed, which have shown excellent performances on the benchmark dataset of SQuAD~\cite{rajpurkar2016squad}.
Recently, the QA models empowered by the pre-trained language models of BERT~\cite{devlin2018bert}, XLNet~\cite{yang2019xlnet}, and RoBERTa~\cite{liu2019roberta} have significantly advanced the performance of the task. Nonetheless, they still cannot effectively handle cases that need complex reasoning. To tackle the problem, several models have been developed. NAQANet~\cite{dua2019drop} adapts the output layer of QANet to numerical reasoning by predicting answers from arithmetic computation over the numbers in a text. NumNet~\cite{ran2019numnet} and QDGAT~\cite{chen2020question} further utilize a numerically-aware graph neural network to encode numbers. The approaches still rely on neural networks, which are good at analogical reasoning but not logical reasoning, and thus cannot fundamentally resolve the problem.

There has also been existing work trying to perform symbolic processing for QA. BERT-Calculator~\cite{andor2019giving} and NeRd~\cite{chen2019neural} generate executable programs to produce the final answers. \cite{gupta2019neural} employ a parser to generate a program comprised of neural modules from the question and a program executor based on neural module networks to find the answer. Unfortunately, the methods are not suitable for all QA problems. There are cases in which deep neural networks based on pre-trained language models can easily make accurate predictions. Employing the symbolic approach alone would not effectively solve the problem.

\paragraph{Natural Language Inference with Logical Reasoning}
NLI is a task of predicting the entailment relation between two texts, i.e., premise and hypothesis. The benchmark datasets for NLI, such as SNLI~\citep{bowman2015large} and MultiNLI~\citep{N18-1101} are widely used. Pre-trained language models achieve state-of-the-art performances~\citep{liu2019roberta}.

However, sometimes logical reasoning is also crucial for NLI~\cite{maccartney2007natural}. It has been shown that using pre-trained language models in NLI is sub-optimal when logical reasoning is needed, when it involves conjunction~\citep{saha2020conjnli} and quantitative reasoning~\citep{ravichander2019equate}. Increasing the size of training data for fine-tuning cannot effectively address the issue. There are also rule-based methods to handle quantitative reasoning~\citep{roy2015reasoning}. However, the results are usually not satisfactory.

\begin{table*}[]
\setlength{\belowcaptionskip}{-0.5cm}
    \centering
    \begin{tabular}{p{4.2cm}p{2.4cm}p{2.7cm}p{2.9cm}p{1.8cm}}
        \toprule
        Premise & Hypothesis & E-Program & C-Program & Symbolic Prediction \\\midrule
      Sam had 98.0{\color{orange}@M1} pennies in his bank and he spent 93.0{\color{orange}@M2} of his pennies. & He has 5.0{\color{orange}@N1} pennies now. & diff(M1, M2)=N1 & diff(M1, M2)!=N1 & Entailment \\
        In a school, there are 542{\color{orange}@M1} girls and 387{\color{orange}@M2} boys. & 928{\color{orange}@N1} pupils are there in that school. & add(M1,M2)=N1 & add(M1,M2)!=N1 & Contradiction \\\bottomrule
    \end{tabular}
    \caption{Two examples of natural language inference requiring numerical reasoning. In our method, the numbers in the input are attached with special tokens (in orange). The E-programs and C-programs are generated for logical reasoning, in addition to analogical reasoning. The symbolic predictions are obtained by executions of the programs.}
    \label{tab:NLI_num_examples}
\end{table*}

\paragraph{Applications of dual-process theory} Several research groups have been working on the application of \emph{System 1} and \emph{System 2} into machine learning. For example, \citep{mittal2017thinking} take the vector space model and reasoning in knowledge graphs as fast thinking and slow thinking, respectively, and propose a hybrid query processing engine for search. \citep{anthony2017thinking} use a tree search as an analog of \emph{System 2} to strengthen the planning in sequential decision-making. \citep{bengio2017consciousness} proposes a consciousness prior theory for learning high-level concepts and points out that the \emph{System 2} abilities are closely related to consciousness. In~\citep{chen2019deep}, the authors propose an end-to-end framework including a generative decoder (fast thinking) and a reasoning module (slow thinking) to solve complex tasks.

\section{Neural-Symbolic Processor}

We describe the proposed framework Neural-Symbolic Processor (NSP) in this section.

\subsection{Overview}
Figure~\ref{framework} shows the architecture of the NSP framework. The framework contains an encoder, several predictors, several decoders and executors, and a mixture-of-experts. There are two types of reasoning: analogical reasoning and logical reasoning. The predictors are for analogical reasoning, and the executors and decoders are for logical reasoning. The encoder and the mixture-of-experts are shared. The framework can be utilized in an NLU task such as QA and NLI.

The encoder takes a pair of texts as input and transforms it into a text embedding (intermediate representation). A predictor takes the embedding as input and generates a neural prediction. The prediction can be classification, span extraction, or sequence tagging. In parallel, a decoder takes the embedding as input and generates a program. The executor executes the program and generates a symbolic prediction. The program can represent a logical reasoning for the task. Finally, the mixture-of-experts takes the neural and symbolic predictions and makes a final prediction. Note that the encoder and each of the decoders form a sequence-to-sequence model. The encoder, predictors, and decoders are all assumed to be based on a pre-trained language model such as BART, RoBERTa, and BERT.

For example, in the first example of QA in Table~\ref{tab:QA_num_examples}, to give the correct answer, one needs to calculate 53-24 to obtain the result of 29. Employing neural reasoning alone would not easily accomplish the task. Using NSP,  we attach 53 and 24 with the special tokens N9 and N10 representing the numbers in the pre-processing. We also generate the program diff(N9, N10) describing the calculation and obtain the correct result by executing the program. This is in parallel with analogical reasoning.

In the first example of NLI in Table~\ref{tab:NLI_num_examples}, to give the correct answer, one needs to calculate 98-93 to obtain the result of 5. In our method, we attach the numbers 98 and 93 with the special tokens M1 and M2. We also generate the two programs and obtain the correct result by executing the programs.

\subsection{Encoder}

The encoder is a Transformer encoder. The output of the encoder is used for both analogical and logical reasoning. (By default there is only one encoder. There can be also two encoders, one for analogical reasoning and the other for logical reasoning.)

There is a pre-processing before the encoder, in which the numbers in the input are attached with special tokens, as shown in Table~\ref{tab:QA_num_examples} and Table~\ref{tab:NLI_num_examples}.

\subsection{Analogical Reasoning}

The system for analogical reasoning predicts the answers from the input, using the predictors. Each of the predictors is a task-specific layer from the encoder. We next give the details in QA and NLI.

\paragraph{QA Task} 
For the QA task, the encoder takes the passage-question pair as input and outputs \texttt{[CLS]} and token representations. In our experiments, the encoder is built on RoBERTa. The predictors take the representations as input and output neural predictions. Each predictor deals with one type of answer: span extraction, sequence labeling, or classification. It is a standard model for QA. 

\begin{itemize}
    \item Span extraction: The predictor predicts the answer as a contiguous span in the passage or in the question. It calculates each token's beginning/end probability and extracts the span with the largest probability. The probability of an answer is defined as the product of the probabilities of the beginning and end tokens.

    \item Sequence labeling: The predictor predicts the answer as non-contiguous spans in the passage using the token representations. It decides for each token whether or not it belongs to the answer.

    \item Classification: The predictor views QA as a classification problem. It predicts the answer as a class label using the \texttt{[CLS]} representation. For example, the classes can be ten digits 0-9. 

\end{itemize}

\paragraph{NLI Task} 

For the NLI task, the encoder takes a pair of texts as input (premise and hypothesis). The predictor
makes a three-class classification using the \texttt{[CLS]} representation, deciding the relation between the text pair: entailment, neutral and contradiction. It is a standard model for NLI.

\subsection{Logical Reasoning}

The system for logical reasoning predicts the answers from the input, using the decoders and executors. A decoder transforms the input into a program based on the output of the encoder. The corresponding executor then executes the program. If the generated program is not valid, the executor will return NULL. In our experiments, the sequence-to-sequence models are based on BART~\cite{lewis2020bart}. Note that the numbers in the input are attached with special tokens for generating the programs, and the programs also utilize the special tokens. The definitions of functions in the programs are given in Appendix~\ref{appendix:program}. We next give details in QA and NLI.

\paragraph{QA Task}

There is only one decoder. The decoder generates the programs for QA, like those in Table~\ref{tab:QA_num_examples}. For example, the first program diff(N9, N10) means subtracting N9 from N10 in the input. The executor takes the program, makes substitutions N9=53 and N10=24, and then obtains the result of 29.

\paragraph{NLI Task}


\begin{table}[]

\setlength{\belowcaptionskip}{-0.5cm}

    \centering
    \begin{tabular}{ccc}
        \toprule
        E-Program & C-Program & NLI prediction \\\midrule
        True & True & Invalid \\
        True & False & Entailment\\
        False & True & Contradiction\\
        False & False & Neutral \\\bottomrule
    \end{tabular}
    \caption{The relation between the outputs of E-Program and C-Program and the NLI predictions.}
    \label{tab:nli-decision}
\end{table}

For the NLI task, one program is insufficient to make a prediction. Therefore, we use two decoders to generate two programs: an entail program (E-Program) and a contradiction program (C-Program). E-Program becomes true if the premise entails the hypothesis. C-Program becomes true if the premise contradicts the hypothesis. Table~\ref{tab:nli-decision} shows how to make the NLI prediction using the results of E-Program and C-Program. 

For example, the E-Program is diff(M1,M2)=N1 and the C-Program is diff(M1,M2)!=N1 in the first example in Table~\ref{tab:NLI_num_examples}.  The first program predicts true if the equation of diff(M1-M2) equals N1 holds, and the second program predicts true if the equation of diff(M1-M2) equals N1 does not hold.
The executors take the programs, make substitutions M1=98.0, M2=93.0, and N1=5.0, obtain the results of true and false, and make the final prediction as entailment, based on the decision rules in Table~\ref{tab:nli-decision}.

\subsection{Mixture-of-Experts}

The mixture-of-experts (MoE) (e.g., ~\cite{shazeer2017outrageously})
takes the neural predictions and symbolic predictions as input and selects the valid prediction with the highest probability (most likely to be correct) given by the gating network. The MoE will not select the logical reasoning result if it is NULL. The gating network is simply a prediction layer on the top of the encoder.

\subsection{Training}

Training has two phases: learning of the neural networks and learning of the MoE.

In the first phase, the encoder, predictors, and decoders are trained using the text inputs, programs, and ground-truth outputs. It is a multi-task learning with three objectives. The first objective is to make accurate predictions from the predictors, the second is to generate correct programs from the decoders, and the third is to make accurate predictions of answer types. The answer types indicate which types the correct answers belong to (classification, span, program execution, etc.).

In the second phase, the gating network of MoE is trained using the text inputs and the probabilities of predictions by answer types. 

In the QA task, there are five types of answers (1) passage span extraction, (2) question span extraction, (3) sequence labeling, (4) number (0-9) classification, and (5) program execution, respectively corresponding to five analogical or logical reasoning modules. For each sample $i$ in the training set, let $q_i=(q_{i1}, \cdots, q_{in})$ be the probabilities of predictions by answer types, where $n$ denotes the number of answer types ($n=5$ here). 
Let $p_i=(p_{i1}, \cdots, p_{in})$ be the outputs of the gating network, also by answer types. 
The parameters of the gating network are learned by using the text inputs and minimizing the KL-divergence in prediction of answer types
\begin{eqnarray}
\mathcal{L} &=& - \sum_{i} \frac{1}{n} \sum_{k=1}^{n} [q_{ik}\log(p_{ik})\nonumber\\
&\quad&+ (1-q_{ik})\log(1-p_{ik})]. \nonumber
\end{eqnarray}


In the NLI task, there are only two types of answers, classification and program execution, respectively. Therefore, we adopt a simple strategy of selecting the type of predictions with the highest accuracy in the development dataset.

\section{Experiment}


\subsection{QA Task}
\subsubsection{Dataset}
DROP~\cite{dua2019drop} is a dataset suitable for question answering requiring complex reasoning, such as multi-span extraction, arithmetic computation, counting, and multi-step reasoning. DROP is constructed from Wikipedia by crowd-sourcing, which contains 77,409 / 9,536 / 9,622 instances in the training / development / testing split.
The training dataset does not have programs that we need in NSP. 
We sample a subset of the training dataset of DROP, named DROP-subset, containing 32,011 instances, and let human annotators annotate programs for the instances. (It is too costly to annotate all DROP data). In addition, we annotate programs for all data in the development dataset. The detailed annotation process is described in Appendix~\ref{annotation}. We use the development and test datasets of DROP as offline test and online test sets, respectively. Note that online test results are obtained from the official website, and thus there is no breakdown of the results.

Following the previous work~\cite{dua2019drop}, we adopt two evaluation metrics, Exact Match (EM) and F1 score, to conduct evaluations.  

\subsubsection{Baselines}
We compare our method with several baselines. The first baseline is NA-RoBERTa (Numerically-Aware RoBERTa), which has a similar architecture to NAQANet~\cite{dua2019drop}, but uses RoBERTa as the encoder. NA-RoBERTa can be regarded as an approximation of using analogical reasoning in our method.

The other baselines are models previously applied to DROP: (1) QANet~\cite{yu2018qanet}, a traditional reading comprehension model combining convolution and self-attention models. (2) NAQANet~\cite{dua2019drop}, a model which improves the output of QANet by adding a module to predict answers through arithmetic computation. (3) NumNet~\cite{ran2019numnet}, a model which uses GNN to enhance the embedding of numerical features. (4) QDGAT~\cite{chen2020question}, a model which improves NumNet by performing GNN on a heterogeneous graph. (5) NeRd~\cite{chen2019neural}, a symbolic reasoning model which uses BERT as encoder and LSTM as a decoder to generate a program and then executes the program to produce the answer. 

\subsubsection{Experimental setting}

For analogical reasoning, the encoder of our method is based on RoBERTa-large. All predictors (span extraction, sequence labeling, number classification) share the same encoder. We perform an end-to-end multi-task training for 20 epochs using the AdamW optimizer~\cite{loshchilov2018fixing} with a batch size of 16. For the encoder, the learning rate is 1.5e-5, and the L2 weight decay is 0.01. For the predictor (prediction layer), the learning rate is 1e-4, and the L2 weight decay is 5e-5.

For logical reasoning, the sequence-to-sequence model of our method is based on BART-large. We train the model for 100 epochs using AdamW, with a batch size of 16. The learning rate is 1e-5. 

\subsubsection{Experimental Results}

Table~\ref{tab:DROP1} shows the experimental results. Our method of NSP outperforms all baselines, achieving 84.01 in F1 score and 80.81 in EM in the offline test. NA-RoBERTa can be regarded as a model only performing neural reasoning. Our method outperforms NA-RoBERTa by 2.40 in F1 score and 2.51 in EM. Our method outperforms QDGAT by 2.60 in terms of F1 score and by 2.56 in terms of EM. Our method also performs better than all baselines in the online test. The experimental results demonstrate the effectiveness of our method. 

\begin{table}[]

\setlength{\belowcaptionskip}{0cm}

\centering

\setlength{\tabcolsep}{1mm}{

\begin{tabular}{lcccc}

\toprule

\multirow{2}{*}{Method} & \multicolumn{2}{c}{Offline Test} & \multicolumn{2}{c}{Online Test} \\
\cline{2-5}
& EM & F1 & EM & F1 \\
\hline
QDGAT           & 78.25 & 81.41 & 78.72 & 82.11   \\
NA-RoBERTa      & 78.30 & 81.61 & 78.76 & 82.25   \\
NSP (our method) & {\bf 80.81} & {\bf 84.01} & {\bf 79.74} & {\bf 83.31}  \\

\bottomrule
\end{tabular}
}
\caption{The results of our method and baselines trained on DROP-subset in offline test and online test.}\label{tab:DROP1}
\end{table}

\begin{table}[]
\setlength{\belowcaptionskip}{-0.5cm}
\centering
\begin{tabular}{lcccc}
\toprule

\multirow{2}{*}{Method} & \multicolumn{2}{c}{Offline Test} & \multicolumn{2}{c}{Online Test} \\
\cline{2-5}
& EM & F1 & EM & F1 \\
\hline

QANet           &  27.50  & 30.44  & 25.50 & 28.36\\
NAQANet         & 46.20   & 49.24 & 44.07 & 47.01\\
NumNet          &  64.92 & 68.31 & 64.56 & 67.97\\
NeRd            & 78.55 & 81.85  & 78.33 & 81.71   \\
QDGAT           & 82.74 & 85.85 & 83.23 & 86.38 \\
\bottomrule
\end{tabular}
\caption{The results of methods trained on the full DROP training dataset in offline test and online test. The results are taken from the original papers.}\label{tab:DROP-others}
\end{table}

Table~\ref{tab:DROP-others} reports the results of existing methods trained on the {\em full} DROP training dataset and evaluated in terms of EM and F1 scores in the same offline test and online test. Except for QDGAT, the results are lower than NSP trained on DROP-subset. Note that when trained on DROP-subset, NSP outperforms QDGAT, as shown in Table~\ref{tab:DROP1}.

We investigate the performances of our method of NSP and the baselines on different answer types in the offline test, as shown in Table~\ref{tab:DROP3}. Our method of NSP performs the best on the answers relating to numbers and dates, demonstrating that NSP is strong in numerical reasoning compared with the other methods.

We further investigate the performances of our method of NSP and the baselines
on different program types, add/diff, max/min, count, and mul/div/avg, as shown in Table~\ref{tab:DROP-sub-group}. We can observe that NSP significantly outperforms in terms of F1 (+56.42) in mul/div/avg. It appears that NSP can generate programs for multiplication, division, and averaging, as shown in Appendix~\ref{DROP-good-case}, while NA-RoBERTa and QDGAT do not have the ability. The F1 of NSP also improves 3.91 and 2.6, respectively, in add/diff and max/min. The improvements are also significant. NSP can generate programs to perform multi-step addition, subtraction, and max/min operations, as shown in Appendix~\ref{DROP-good-case}. In contrast, it is hard for NA-RoBERTa and QDGAT to do so. The percentage of generated programs with invalid execution results (NULL) on the DROP-dev dataset is 0.1\%.

\setlength{\tabcolsep}{0.8mm}{
\begin{table}[]
\centering
\begin{tabular}{lcccc}
\toprule

Method & Number & Span(s)  & Date & Total \\
\hline
QDGAT           & 80.41  & 83.90 & 65.38 & 81.41   \\
NA-RoBERTa      & 80.54   & 84.14 & 67.36 & 81.61   \\
NSP (our method) & {\bf 84.24}  & {\bf 84.36} & {\bf 68.49} & {\bf 84.01}  \\

\bottomrule
\end{tabular}
\caption{The results of our method and baselines trained on DROP-subset and evaluated in terms of F1 on offline test. Number, span(s) and date are three answer types.}\label{tab:DROP3}
\end{table}
}

\setlength{\tabcolsep}{1mm}{
\begin{table}[]
\setlength{\belowcaptionskip}{-0.5cm}
\centering
\begin{tabular}{p{2.8cm}p{1cm}p{1cm}p{1cm}p{1cm}}
\toprule

Method & add/ diff & max/ min & count &  mul/ div/avg \\
\hline
Number of cases & 4317 & 282 & 913 & 52 \\
\hline
QDGAT & 81.99 & 92.15 & 79.85 & 4.48 \\
NA-RoBERTa  & 81.49   & 92.68 & 82.15  & 8.33  \\
NSP (our method) & {\bf 85.90}  & {\bf 95.28} & {\bf 82.91}  & {\bf 64.75} \\

\bottomrule
\end{tabular}
\caption{The results of our method and baselines in terms of F1 on cases corresponding to different program types, add/diff, max/min, count, and mul/div/avg.}\label{tab:DROP-sub-group}
\end{table}
}

\subsection{Ablation Study}

We conduct an ablation study. Table~\ref{DROP-ablation} presents the results. We compare NSP, and its components of logical reasoning only and analogical reasoning only in offline test with answer-types of number, span(s), and date. The results indicate that NSP performs much better than the two components, indicating that both components are necessary for NSP. Analogical reasoning is suitable for the span(s)-type. On the other hand, logical reasoning is suitable for the number and date types.

\setlength{\tabcolsep}{0.8mm}{
\begin{table}
\centering
\begin{tabular}{p{3cm}cccc}
\toprule

Method & Number & Span(s) & Date  & Total \\
\hline
Logical reasoning only &  83.23  &  5.62 & 52.08 & 54.58\\
Analogical reasoning only & 27.49  & {\bf 84.39} & 63.87 &  48.74  \\
NSP (our method) & {\bf 84.24}  &  84.36 & {\bf 68.49} & {\bf 84.01}  \\
\bottomrule
\end{tabular}
\caption{Ablation study results on offline test of DROP. Compare only ensemble logical reasoning module and only ensemble analogical reasoning module with NSP. }\label{DROP-ablation}
\end{table}
}

\setlength{\tabcolsep}{0.8mm}{
\begin{table}[]
\setlength{\belowcaptionskip}{-0.5cm}

\centering
\begin{tabular}{p{2.8cm}cccc}
\toprule

Method & Number & Span(s) & Date  & Total \\
\hline
passage span extraction & 15.52  & 78.32 & 63.87 &  39.11  \\
question span extraction & 0.24  & 38.18 & 26.61 & 14.44 \\
sequence labeling &  5.47 & 58.91 & 41.07 &  25.45\\
classification (0-9) & 20.19 & 0.86 & 0.30 & 12.85 \\
program  &  83.23  &  5.62 & 52.08 & 54.58\\
NSP (our method) & {\bf 84.24}  & {\bf 84.36} & {\bf 68.49} & {\bf 84.01}  \\
\bottomrule
\end{tabular}
\caption{The results of NSP and its components in terms of F1 score.}\label{tab:DROP2}
\end{table}
}

Table~\ref{tab:DROP2} reports the F1 scores of predictors and decoder of NSP for different answer types. We observe that the F1 score of NSP, after combining all components, achieves the highest performances in the number, span(s), and date types. It indicates that the components are complementary and that NSP can successfully ensemble them.

NSP utilizes additional program annotations compared with the baselines above. For further comparison, we consider two other alternatives based on NA-RoBERTa that also use the program annotations. The first method adds a program generation sub-task during the training, and only uses the original neural predictions in the inference. The second method utilizes programs to supervise its arithmetic computation, which predicts each number's coefficients (+1, -1, 0). In Table~\ref{baseline_with_program}, we report the F1 scores of the additional methods. We observe that NSP still works the best. NA-RoBERTa with program supervision outperforms NA-RoBERTa. However, NA-RoBERTa with program generation performs even worse than NA-RoBERTa. 

For the MoE in NSP, we train a gating network using text as input. We compare the MoE method with two alternatives, which respectively take the probabilities of answer predictions and the probabilities of answer-type predictions as input of the gating network. Table~\ref{MoE-analysis} shows the results. We observe that using text as input performs the best.

\subsection{NLI Task}

\subsubsection{Dataset}

AWPNLI~\citep{ravichander2019equate} is a dataset suitable for NLI requiring reasoning. AWPNLI is created from arithmetic math problems in which symbolic processing is needed. AWPNLI is a small dataset with only 722 instances. We have human annotators annotate the E-Program and C-Program for each instance. The annotation process is described in Appendix~\ref{annotation}.

\setlength{\tabcolsep}{0.8mm}{
\begin{table}[]
\centering
\begin{tabular}{p{3.2cm}cccc}
\toprule
Method & Number & Span(s) & Date & Total \\
\hline
NA-RoBERTa & 80.54 & 84.14 & 67.36 & 81.61 \\
NA-RoBERTa w/ program generation & 79.43 & 82.90 & 66.37 & 80.46 \\
NA-RoBERTa w/ program supervision  & 81.70 & 84.18 & 67.55 & 82.35 \\
NSP (our method) & {\bf 84.24} & {\bf 84.36} & {\bf 68.49} & {\bf 84.01} \\
\bottomrule
\end{tabular}
\caption{Results of different baseline methods utilizing program annotation.}\label{baseline_with_program}
\end{table}
}

\setlength{\tabcolsep}{0.8mm}{
\begin{table}[]
\setlength{\belowcaptionskip}{-0.5cm}
\centering
\begin{tabular}{lcccc}
\toprule
Input of GN & Number & Span(s) & Date & Total \\
\hline
answer-prob & 82.31 & 82.13 & 68.39 & 82.00 \\
answer-type-prob & 82.96 & 83.62 & {\bf 69.69} & 82.97 \\
text & {\bf 84.24} & {\bf 84.36} & 68.49 & {\bf 84.01} \\
\bottomrule
\end{tabular}
\caption{Results of different input of gating networks (GN) in NSP.}\label{MoE-analysis}
\end{table}
}

\subsubsection{Baselines}
We compare our method with neural and rule-based baselines. For the neural baselines, we use RoBERTa and BART as models for classification. 
The rule-based method Q-REAS proposed 
in~\citep{ravichander2019equate} is also chosen. The performances of Q-REAS and two neural models are taken from the original paper~\citep{ravichander2019equate}.

RoBERTa and BART can be viewed as NSP with only analogical reasoning. We also consider NSP with only logical reasoning (also based on BART). The baselines can serve for ablation study.

\subsubsection{Experimental Setting}
We conduct a ten-fold cross-validation on the AWPNLI dataset and report the average and standard deviation of the results. The sequence-to-sequence model of our method is fine-tuned by using the input text and the annotated program based on BART. The predictor model is fine-tuned by using the input text and the ground truth based on RoBERTa. The optimizer is Adam, and the learning rate is 1e-5. The learning converges in 25k steps.

\subsubsection{Results and Discussion}

Table~\ref{tab:nli} shows the results of methods on the AWPNLI dataset. 
We find that our method of NSP is slightly better than its variant of logical reasoning only. The accuracies of the two methods are significantly better than those of the other methods. The results indicate that our method of NSP is effective when numerical reasoning is needed in NLI.

Furthermore, we can see that the methods of RoBERTa and BART are much worse than NSP and logical reasoning only. 
The results reported in previous work under the zero-shot setting are also significantly lower. We conclude that utilization of local reasoning capabilities such that of NSP is necessary for the task.

\begin{table}[t]
\setlength{\belowcaptionskip}{-0.5cm}
\centering
\begin{tabular}{lc}
\toprule
Model & Accuracy              \\\midrule
\multicolumn{2}{l}{\emph{Zero Shot Evaluation}}     \\
BART-large             & 42.2*                \\
GPT                    & 50.0*                \\
Rule-based Q-REAS             & 71.5*                \\\midrule
\multicolumn{2}{l}{\emph{10-fold Cross Validation}} \\
RoBERTa & 49.85±0.35\\
BART & 49.85±6.28\\
Logical Reasoning Only      & \textbf{88.05±4.71} \\
NSP (our method)        & \textbf{92.24±4.68} \\       
\bottomrule
\end{tabular}
\caption{The results of NSP and baselines on AWPNLI. The results with * are taken from the original papers.}\label{tab:nli}
\end{table}

\section{Conclusion}

This paper proposes a novel framework for natural language understanding (NLU), referred to as Neural-Symbolic Processor (NSP). NSP employs two types of reasoning, analogical reasoning and logical reasoning. To `understand' language, analogical reasoning is performed by using neural networks as usual. In the meantime, logical reasoning is performed by using neural networks to generate programs and then using symbolic systems to execute the programs. Such an architecture is similar to that of humans, and the two types of reasoning correspond to \emph{System 1} and \emph{System 2} respectively in the human brain. Our approach thus is powerful in dealing with the challenging problems which conventional neural-network-only approaches suffer from. We evaluate our approach in two NLU tasks, QA and NLI. The experiments show that our method surpasses previous state-of-the-art methods with remarkable improvements when logical reasoning is needed.

\section*{Limitations}

Although NSP outperforms the baselines in both QA and NLI, there are still complicated cases that the current NSP cannot effectively deal with. 

For the QA task, We conduct an error analysis of NSP on the DROP dataset. Table~\ref{DROP-bad-example} provides three typical categories of hard cases for NSP. The first type (first example) is related to multi-step reasoning. Such cases need a deep reasoning path or many arguments in functions. Another type (second example) is about the counting of long strings. Defining a standard for program annotation is hard because different human annotators may select strings with different lengths. The last type (third example) is related to complex conditions. The programs of NSP currently use a simple grammar language, which still cannot represent complicated conditions in reasoning. 

We do not introduce a complicated programming language, because there is not enough training data to learn a model for generating complex programs. We leave this to future work. 

\begin{table*}[]
    \centering
    \begin{tabular}{p{9cm}p{6cm}c}
    \toprule
    Passage \& Question & Prediction \\\midrule
    \emph{Complicated multi-step computation} \\\midrule
    {\bf Passage:} ... In the second quarter@N4, New Orleans regained the lead as QB Drew Brees ( a former Charger ) completed a 12 - yard@N5 TD pass to WR Devery Henderson ( with a failed PAT ) and RB Deuce McAllister getting a 1 - yard@N6 TD run. San Diego answered as QB Philip Rivers completed a 12 - yard@N7 TD pass to RB LaDainian Tomlinson, but the Saints replied with Brees completing a 30 - yard@N8 TD pass to WR Lance Moore. The Chargers closed out the half with Rivers completing a 12 - yard@N9 TD pass to TE Antonio Gates. In the third quarter@N10, New Orleans increased its lead Brees completing a 1 - yard@N11 TD pass to TE Mark Campbell ... San Diego tried to rally as Kaeding nailed a 31 - yard@N16 field goal, Rivers completed a 14 - yard@N17 TD pass to WR Vincent Jackson \newline {\bf Question:} How many more yards of touchdown passes did Drew Brees make than Philip Rivers? & {\bf Prediction}: diff(add(N5,N8,N11,N17),add( N7,N9)) \newline {\bf Result:} 33 \newline {\bf Manual:} diff(add(N5,N11,N8),add(N7,N9,N17)) \newline {\bf Ground truth:} 5 \\\midrule
    \emph{Counting long strings}\\\midrule
    {\bf Passage:} ... They explain that the gap may persist due to the crack epidemic, the degradation of African - American family structure, the rise of fraud in the educational system ( especially with respect to No Child Left Behind ), the decrease in unskilled real wages and employment among African - Americans due to globalization and minimum wage increases, differences in parental practices ( such as breastfeeding or reading to children ), and " environmental conditions shaped by [ African - Americans ] themselves. ... \newline {\bf Question:} How many reasons are cited as causing the persistent gap between white and black IQs? & {\bf Prediction}: count("crack epidemic","the degradation of African - American family structure","the rise of fraud in the educational system")"  \newline {\bf Result:} 3 \newline {\bf Manual:} count("crack epidemic","the degradation of African - American family structure","the rise of fraud in the educational system ( especially with respect to No Child Left Behind )","the decrease in unskilled real wages and employment among African - Americans due to globalization and minimum wage increases","differences in parental practices ( such as breastfeeding or reading to children )","environmental conditions shaped by [ African - Americans ] themselves") \newline {\bf Ground truth:} 6 \\\midrule
    \emph{Questions with complex conditions}\\\midrule
    {\bf Passage:} ... yet the Raiders would answer with kicker Sebastian Janikowski getting a 33 - yard@N5 and a 30 - yard@N6 field goal. Houston would tie the game in the second quarter@N7 with kicker Kris Brown getting a 53 - yard@N8 and a 24 - yard@N9 field goal. ... \newline {\bf Question:} How many field goals did both teams kick in the first@Q0 half ? & {\bf Prediction}: count(N5,N6,N8,N9) \newline {\bf Result:} 4 \newline {\bf Manual:} count(N5,N6) \newline {\bf Ground truth:} 2 \\\midrule
    
    \end{tabular}
    \caption{Examples of incorrect predictions by NSP in offline test of DROP.}\label{DROP-bad-example}
\end{table*}

We also conduct an error analysis of NSP for the NLI task. The primary type of errors (hard cases) is related to redundant numbers in the input texts, as shown in Table~\ref{NLI-bad-example}. One possible solution would be to increase the size of training data.

\begin{table*}[]
    \centering
    \begin{tabular}{p{5.8cm}p{4.7cm}p{2.1cm}p{2.1cm}}
    \toprule
    Premise \& Hypothesis & Program & Result & Ground Truth \\\midrule
    \emph{Disturbance by redundancy numbers}\\\midrule
    {\bf Premise:} Melanie picked 7.0@M1 plums and 4.0@M2 oranges from the orchard and Sam gave her 3.0@M3 plums. \newline
    {\bf Hypothesis:} She has 10.0@N1 plums now. & {\bf Prediction:} \newline
    E: add(diff(M1,M2),M3)=N1 \newline
    C: add(M1,M3)!=N1 \newline
    {\bf Manual:} \newline
    E: add(M1,M3)=N1 \newline
    C: add(M1,M3)!=N1 & neutral & entailment\\\midrule
    {\bf Premise:} Sally had 39.0@M1 baseball cards , and 9.0@M2 were torn and Sara bought 24.0@M3 of Sally 's baseball cards.\newline
    {\bf Hypothesis:} Sally has 15.0@N1 baseball cards now. & {\bf Prediction:} \newline
    E: diff(add(M1,M2),M3)=N1 \newline
    C: diff(add(M1,M2),M3)!=N1 \newline
    {\bf Manual:} \newline
    E: diff(M1,M3)=N1 \newline
    C: diff(M1,M3)!=M1 & contradiction & entailment \\\midrule
    \end{tabular}
    \caption{Examples of incorrect predictions by NSP on AWPNLI.}\label{NLI-bad-example}
\end{table*}

\section*{Acknowledgements}
The authors thank Yuchen Zhang at ByteDance for his insightful comments in technical discussion. The authors thank anonymous reviewers for their helpful suggestions.

\begin{table*}[]
    \centering
    \begin{tabular}{p{1.5cm}p{5cm}p{7.5cm}}
        \toprule
        Function & Arguments & Description \\\midrule
        add & a set of special tokens/programs & compute the summation of the terms\\\midrule
        diff & a pair of special tokens/programs & compute the difference of the two terms \\\midrule
        max \newline min & a set of special tokens/programs & select the maximum/minimum among the terms \\\midrule
        mul \newline div & a pair of special tokens/programs & compute the multiplication/division of the two terms\\\midrule
        avg & a set of special tokens/programs & compute the average of the terms\\\midrule
        count & a set of special tokens/programs & count the number of the terms\\\midrule
        year \newline month \newline day & a span in input text & convert the span to the corresponding year/month/day in numerical expression  \\
        \midrule
        hour \newline minute \newline second & a span in input text & convert the span to the corresponding hour/minute/second in numerical expression  \\
        \midrule
        = \newline != & a pair of special tokens/programs & return two terms are equal/not equal\\
        \midrule
    \end{tabular}
    \caption{Description of the functions in the programs.}
    \label{function-definition}
\end{table*}

\bibliography{anthology,custom}
\bibliographystyle{acl_natbib}

\appendix

\section{Definition of functions in programs}\label{appendix:program}
The functions in programs are provided in Table~\ref{function-definition}.

\section{Annotation Process}\label{annotation}

The annotation team has ten annotators from a commercial company in data annotation. We sign a contract with the company and pay the company for the annotation work at a market price in China. They are all college graduates with high capabilities in English and math. The required skill is the capability of understanding the texts in English and formulating the answering processes into programs. Both the QA and NLI datasets have the same annotation process. The annotation task is that given input texts with special tokens and the corresponding labels, the annotator: a) decides whether it needs logical reasoning to get the answers; b) if yes, labels the corresponding programs; if not, labels a [NULL]. We first conduct training for the annotators about the annotation rules. Then we separate the dataset into ten batches. Each annotator labels one batch of the dataset. We check 15\% of samples in each batch. The whole batch will be relabeled if annotation accuracy is below 90\%. Each data instance is labeled by only one annotator.

\section{Examples only NSP can predict correct results}\label{DROP-good-case}
Table~\ref{DROP-good-example} gives examples that only NSP can give correct results. 

\begin{table*}[]
    \centering
    \begin{tabular}{p{7cm}p{4.5cm}p{3.5cm}c}
        \toprule
        passage & question & prediction \\\midrule
        \emph{mul \& div \& avg operators} \\\midrule
        ... The Dolphins ended the period with kicker Jay Feely getting a 53 - yard@N6 field goal. In the second quarter@N7, Miami drew closer as Feely kicked a 44 - yard@N8 field goal, yet New York replied with kicker Mike Nugent getting a 29 - yard@N9 field goal. ... & How many yards long does the average field goal measure when only the first@Q0 three@Q1 are taken into account ? & {\bf Prediction}: div(add(N6,N8,N9),Q1) \newline {\bf Result:} 42 \newline {\bf Ground truth:} 42\\
        \midrule
        ... Dave Rayner nailed a 23-yard@N5 field goal ... Green Bay managed to get two@N8 more field goals , as Rayner got a 54-yarder@N9 and a 46-yarder@N10 to end the half ... David Akers got a 40-yard@N13 field goal ... & How many yards long was the average length across all field goals scored ? & {\bf Prediction}: avg(N5,N9,N10,N13) \newline {\bf Result:} 40.75 \newline {\bf Ground truth:} 40.75\\\midrule
        \emph{add \& diff operators} \\\midrule
        ... Jermichael Finley caught a 20-yard@N6 touchdown pass from Aaron Rodgers ... The Lions responded with a Calvin Johnson 25-yard@N10 touchdown pass from Matthew Stafford ... The Packers then scored a touchdown when Randall Cobb caught a 22-yard@N13 pass from Aaron Rodgers ... & How many more yards of touchdown passes did Aaron Rodgers throw than Matthew Stafford ? & {\bf Prediction}: diff(add(N6,N13),N10) \newline {\bf Result:} 17 \newline {\bf Ground truth:} 17\\
        \midrule
        ... Of 162@N1 cities worldwide, MasterCard ranked Bangkok as the top destination city by international visitor arrivals in its Global Destination Cities Index 2018@N2 ... & How many cities were n't ranked the top destination city by international visitor arrivals in the Global Destination Cities Index 2018@Q0 & {\bf Prediction}: diff(N1,\newline count("Bangkok")) \newline {\bf Result:} 161 \newline {\bf Ground truth:} 161\\
        \bottomrule
    \end{tabular}
    \caption{Examples of questions related to numerical reasoning on DROP, where only NSP can give correct program and final results.}\label{DROP-good-example}
\end{table*}

\end{document}